\def\year{2018}\relax
\documentclass[letterpaper,table]{article} 
\usepackage{aaai18}  
\usepackage{times}  
\usepackage{helvet}  
\usepackage{courier}  
\usepackage{url}  
\usepackage{graphicx}  
\usepackage{booktabs}
\usepackage{wrapfig}
\usepackage{xcolor}
\usepackage{amsmath}
\usepackage{stackengine}
\usepackage{array}
\usepackage{subcaption}
\usepackage{multirow}

\definecolor{lgray}{gray}{0.9}
\newcommand\citet[1]{\citeauthor{#1} \shortcite{#1}}
\begin{document}
\title{Learning Interpretable Spatial Operations in a Rich 3D Blocks World}
\author{Yonatan Bisk$^{1*}$\ \ \ \ \ \ Kevin J. Shih$^2$\ \ \ \ \ \ Yejin Choi$^1$\ \ \ \ \ \ Daniel Marcu$^3$\thanks{Work performed at USC's Information Sciences Institute.}\\
$^1$Paul G. Allen School of Computer Science \& Engineering, University of Washington\\ 
$^2$University of Illinois at Urbana-Champaign\ \ \ \ \ \ \ $^3$Amazon Inc.\\ 
{ \url{{ybisk,yejin}@cs.washington.edu}, \ \ \ \  \url{kjshih2@illinois.edu}, \ \ \ \
\url{marcud@amazon.com}
  }
}
\maketitle
\begin{abstract}
In this paper, we study the problem of mapping natural language instructions to complex spatial actions in a 3D blocks world. 
We first introduce a new dataset that pairs complex 3D spatial operations to rich natural language descriptions that require complex spatial and pragmatic interpretations such as \emph{``mirroring'', ``twisting''}, and  \emph{``balancing''}. 
This dataset, built on the simulation environment of \citet{Bisk:2016:NAACL}, attains language that is significantly richer and more complex, while also doubling the size of the original dataset in the 2D environment with 100 new world configurations and 250,000 tokens. 
In addition, we propose a new neural architecture that achieves competitive results while automatically discovering an inventory of interpretable spatial operations (Figure \ref{fig:operations_grid}).

\end{abstract}

\section{Motivation}

One of the longstanding challenges of AI, first introduced as SHRDLU in early 70s \cite{shrdlu}, is to build an agent that can follow natural language instructions in a physical environment. 
The ultimate goal is to create systems that can interact in the \textit{real} world using \emph{rich} natural language. 
However, due to the complex interdisciplinary nature of the challenge \cite{harnard}, which spans across several fields in AI, including robotics, language, and vision, most existing studies make varying degrees of simplifying assumptions. 

On one end of the spectrum is rich robotics paired with 
simple constrained language \cite{Roy:2005fj,tellex:2011}, as 
acquiring a large corpus of natural language grounded with a real robot is prohibitively expensive \cite{Misra-RSS-14,thomason:corl17}.  
On the other end of the spectrum are approaches based on simulation environments, 
which support broader deployment at the cost of unrealistic simplifying assumptions about the world 
\cite{Bisk:2016:NAACL,wang-liang-manning:2016:P16-1}.
In this paper, we seek to reduce the gap between two complementary research efforts by introducing a new level of complexity to both the environment and the language associated with the interactions. 

\begin{figure}
\centering
\begin{tabular}{c@{\hspace{2pt}}c@{\hspace{2pt}}c}
\multicolumn{3}{c}{\it ``On the (new) fourth tower, mirror Nvidia with UPS.''}\\[3pt]
\raisebox{-.5\height}{\includegraphics[width=0.4\linewidth]{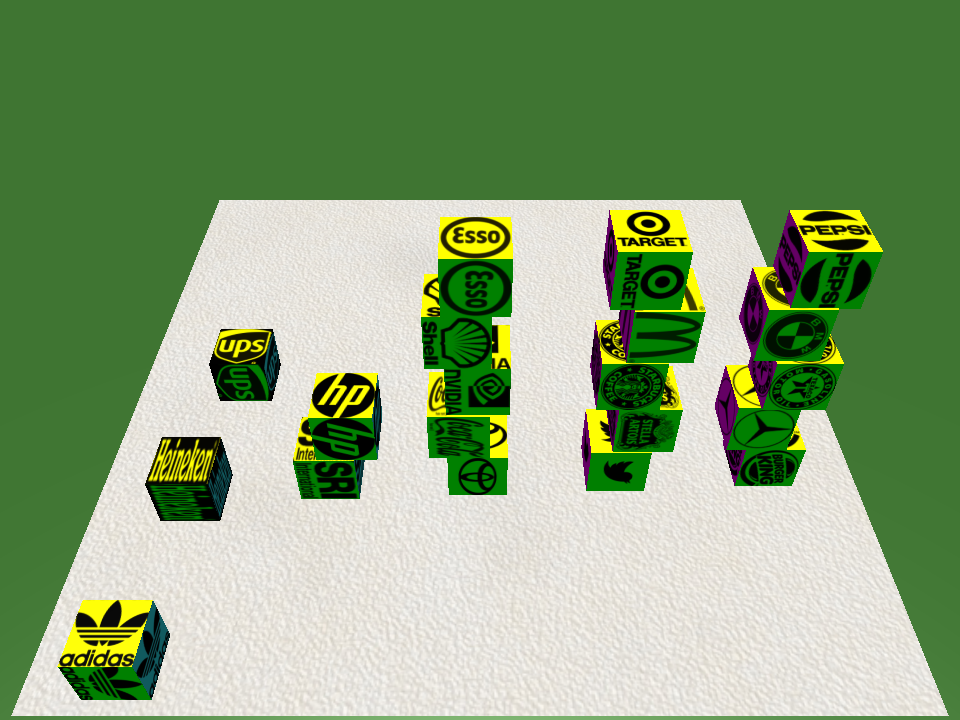}} & $\Rightarrow$ & \raisebox{-.5\height}{\includegraphics[width=0.4\linewidth]{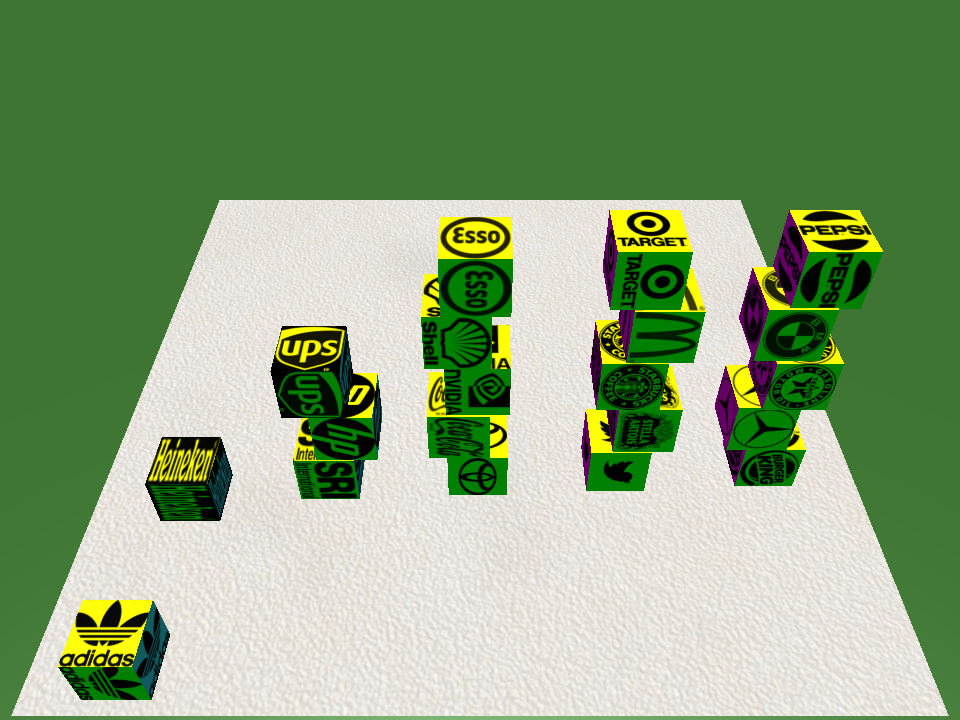}}\\
\end{tabular}
\caption{Example language instruction in our new dataset.  The action requires fine-grained positioning and utilizes a complex concept: {\it mirror}.}
\label{fig:example}
\end{figure}

\paragraph{Lifting Grid Assumptions}
We find that language situated in a richer world leads to richer language.  One such example is presented in Figure \ref{fig:example}.
To correctly place the UPS block, the system must understand the complex physical, spatial, and pragmatic meaning of language including: 
(1) the 3D concept of a \emph{tower}, 
(2) that \textit{new} or \textit{fourth} are referencing an assumed future,
and (3) that \textit{mirror} implies an axis and reflection.  
However, concepts such as above are often outside the scope of most  existing language grounding systems.

In this work, we introduce a new dataset that allows for learning significantly richer and more complex spatial language than previously explored. 
Building on the simulator provided by  \citet{Bisk:2016:NAACL}, 
we create roughly 13,000 new crowdsourced instructions (9 per action), nearly doubling the size of the original dataset in the 2D blocks world introduced in their previous work. 
We address the challenge of realism in the simulated data by introducing three crucial but previously absent complexities:
\begin{enumerate}
\item 3D block structures (lifting 2D assumptions)
\item Fine-grained real valued locations (lifting grid assumptions)
\item Rotational, angled movements (lifting grid assumptions)
\end{enumerate}

\begin{figure*}[h]
\centering
\begin{tabular}{@{}c@{\hspace{2pt}}c@{\hspace{30pt}}c@{\hspace{2pt}}c@{}}
\multicolumn{2}{c}{\textbf{\citet{Bisk:2016:NAACL}}} & \multicolumn{2}{c}{\textbf{This work}}\\
\includegraphics[width=0.2\linewidth]{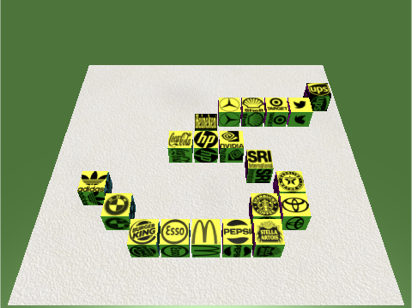} &
\includegraphics[width=0.2\linewidth]{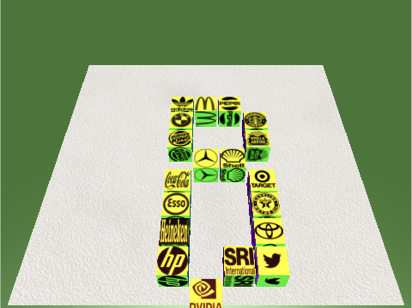} &
\includegraphics[width=0.2\linewidth]{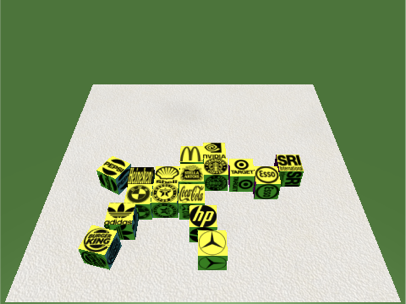} &
\includegraphics[width=0.2\linewidth]{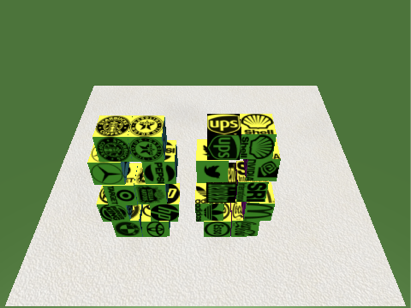} \\
\multicolumn{2}{@{}l}{\textbf{Freq Relations: } left, up, right, directly, above, until} & 
\multicolumn{2}{@{}l}{\textbf{New Relations: } degrees, rotate, clockwise, covering,}\\
\multicolumn{2}{@{}l}{corner, top, down, below, bottom, slide, space, between} & 
\multicolumn{2}{@{}l}{45, layer, mirror, arch, towers, equally, twist, balance, ...}
\end{tabular}
\caption{Example goal states in our work as compared to the previous Blocks dataset.  Our work extends theirs to include rotations, 3D construction, and human created designs.  This has a dramatic effect on the language used.  Rich worlds facilitate rich language, above are the most common relations in their data and the most common new relations in ours.}
\label{fig:comparison}
\end{figure*}

\paragraph{Learning Interpretable Operators}
In addition, we introduce an interpretable neural model for learning spatial operations in the rich 3D blocks world. 
In particular, in our model instead of using a single layer conditioned on the language for interpreting the operations, we have the model choose which parameters to apply via a softmax over the possible parameter vectors to use.
Specifically, by having the model decide for each example which parameters to use, the model picks among 32 different networks, deciding which is appropriate for a given sentence.  Learning these networks and when to apply them enables the model to cluster spatial functions.  Secondly,
by encouraging low entropy in the selector, the model converges to nearly one-hot representations during training.  A side effect of this decision is that the final model exposes an API which
can be used interactively for focusing the model's attention and choosing its actions.  
We will exploit this property when generating plots in Figure \ref{fig:operations_grid} showing the meaning of each learned function.
Our model is still fully end-to-end trainable despite choosing its own parameters and composeable structure, leading 
to a modular network structure similar to \cite{andreas-EtAl:2016:N16-1}.

The rest of the paper is organized as follows. We first discuss related work, introduce our new dataset, followed by our new model. We then present empirical evaluations, analysis on the internal representations, and error analysis. We conclude with the discussion for future work. 

\section{Related Work} \label{sec:related}
Advances in robotics, language, and vision are all applicable to this domain. 
The intersection of robotics and language have seen impressive results in 
grounding visual attributes \cite{Kollar:2013wr,Matuszek:2014ts}, spatial reasoning \cite{Steels:1997we,Roy:2002jc,Guadarrama:2013tr}, and action taking \cite{MacMahon:2006:WTC:1597348.1597423,yu-siskind:2013:ACL2013}.  For example, recent work \cite{thomason:ijcai15,thomason:ijcai16,thomason:corl17} has shown how these instructions can be combined with exploration on physical robotics to follow instructions and learn representations online.  

Within computer vision Visual Question Answering \cite{VQA} has been widely popular.  Unfortunately, it is unclear what models are learning and how much they are understanding versus memorizing bias in the training data \cite{anchor:nipsws16}.  Datasets and models have also recently been introduced for visual reasoning \cite{johnson2017inferring,NIPS2017_7082} and referring expressions \cite{kazemzadeh2014referitgame,mao2016generation}.

Finally, within the language community, interest in action understanding follows naturally from research in semantic parsing \cite{andreas-klein:2015:EMNLP,Artzi:2013th}.  Here, the community has traditionally been focused on more complex and naturally occurring text, though this has not always been possible for the navigation domain.

Simultaneously, work within NLP \cite{Bisk:2016:AAAI,wang-liang-manning:2016:P16-1} and Robotics \cite{li2016spatial} returned to the question of action taking and scene understanding in SHRDLU style worlds.  The goal with this modern incarnation was to truly solicit natural language from humans without limiting their vocabulary or referents.  This was an important step in moving towards unconstrained language understanding.

The largest corpus was provided by \citet{Bisk:2016:NAACL}.  In this work, the authors presented pairs of scenes
with  simulated blocks to users of Amazon's
Mechanical Turk.  Turkers would then describe actions or instructions that
their imagined collaborator needs to perform to transform the input scene into the target (e.g. Moving a block to the side of another).  An important aspect of
this dataset is that participants
assume they are speaking to another human.  This means they do not
limit their vocabulary, space of references, simplify their grammar,
or even write carefully.  The annotators assume that whomever will be reading
what they submit is capable of error correction, spatial reasoning,
and complex language understanding.  This provides an important, and
realistic, basis for training artificial language understanding
agents.  Follow-up work has investigated advances to language
representations \cite{pivsl-marevcek:2017:RoboNLP}, spatial
reasoning \cite{tan17}, and reinforcement learning approaches for
mapping language to latent action sequences \cite{misra2017}.  

\section{Creating Realistic Data}
\label{sec:data}
To facilitate closing the gap between simulation and reality, blocks should not have perfect locations, orderings, or alignments.  They should have jitter, nuanced alignments, rotations and the haphazard construction of real objects.  Figure \ref{fig:comparison} shows example how our new configurations aim to capture that realism (right) as compared to previous work (left).  Previous work created target configurations by downsampling MNIST \cite{lecun-98} digits. This enabled them to create interpretable but unrealistic 2D final representations and the order in which blocks were combined was determined by a heuristic to simulate drawing/writing.

In our data, we solicited creations from people around our lab and their children, not affiliated with the project.  They built whatever they wanted (open concept domain), in three dimensions, and were allowed to rotate the blocks.  For example, the animal on the left is an elephant whose trunk, tail, and legs curve.  Additionally, because humans built the configurations, we were able to capture the order in which blocks were placed for a more natural trajectory.  Realism brings with it important new challenges discussed below.

\paragraph{Real Valued Coordinate Spaces}
The discretized world seen in several recent
datasets \cite{Bisk:2016:NAACL,wang-liang-manning:2016:P16-1} simplifies 
spatial reasoning. Simple constructions like \texttt{left} and \texttt{right}
can be reduced to exact offsets that do not require context specific interpretations
(e.g. \texttt{right} $= +[1,0,0]$).  In reality, these concepts
depend on the scene around them.  For example, in the rightmost image
of Figure \ref{fig:comparison}, it is correct to say that the
McDonald's block is \textit{right} of Adidas, but also that SRI
is \textit{right} of Heineken, despite both having different
offsets. The modifier \textit{mirroring} disambiguates the meaning for us. 

\paragraph{Semantically Irrelevant Noise}
It is important to note that with realism comes noise.  Occasionally, an annotator may bump a
block or shift the scene a little. Despite repeated efforts to 
clean and curate the data, most
people did not consider this noise noteworthy because it was
semantically irrelevant to the task.  For example, if while performing
an action, a nearby block jostles, it does not change the holistic 
understanding of the scene.  For this reason, we only evaluate the
placement of the block that ``moved the furthest.''  This is a baby
step towards building models invariant to changes 
in the scene orthogonal to the goal.

\paragraph{Physics}
One concession we were forced to make was
relaxing physics.  Unlike prior
work \cite{wang-EtAl:2017:Long3}, we insisted that the final
configurations roughly adhere to physics (e.g. minimizing overhangs,
no floating blocks, limited intersection), but we found volunteers too
often gave up if we forced them to build entirely with physics turned
on.  This also means that intermediary steps that in the real world
require a counter-weight can be constructed one step at a time.  

\begin{table}
\centering
\begin{tabular}{@{}lllllc@{}}
       & Configs & Types & Tokens & Utters & Ave. Len\\
\toprule
B16 & 100 & 1,281 & 258,013 & 16,767 & 15.4\\ This & 100 & 1,820 &
233,544 & 12,975 & 18.0\\
\midrule        
Total & 200 & 2,299 & 491,557 & 29,742 & 16.5\\
\bottomrule
\end{tabular}
\caption{Corpus statistics for our dataset as compared to previous work (Bisk 16), and the total statistics when combined.}
\label{tab:corpus_stats}
\end{table}

\paragraph{Language}
Our new corpus contains nearly all of the concepts of previous work, but introduces many more.   Figure \ref{fig:comparison} shows the most common 
relations in prior work, and the most common new concepts.  We see that these
predominantly focus on rotation (\textit{degrees, clockwise, ...}) and 3D construction (\textit{arch, balance, ...}), but higher level concepts like mirroring or balancing pose fundamentally new challenges. 

\subsection{Corpus Statistics}
Our new dataset comprises 100 configurations split 70-20-10
between training, testing, and development.  Each configuration has
between five and twenty steps (and blocks).  We present type and token
statistics in Table \ref{tab:corpus_stats}, where we use
NLTK's \cite{NLTK} treebank tokenizer.  This yields higher
token counts in previous works due to
different assumptions about punctuation.

Not all of our annotators
made use of the full 20 blocks.  As such, we have fewer
utterances than the original dataset for the same number of goal
configurations.  Yet, we find that the instructions for
completing our tasks are more nuanced and therefore result in slightly
longer sentences on average.  Finally, we note that while the
datasets are similar, there are significant enough differences that one
should not simply assume that training on the combined dataset will
necessarily yield a ``better" model on either one individually.  There
are important linguistic and spatial reasoning differences between the
two that make our proposed data much more difficult. We present
all modeling results on both subsets of the data and the full combined
dataset.

\begin{figure*}
\centering
\includegraphics[width=0.65\linewidth]{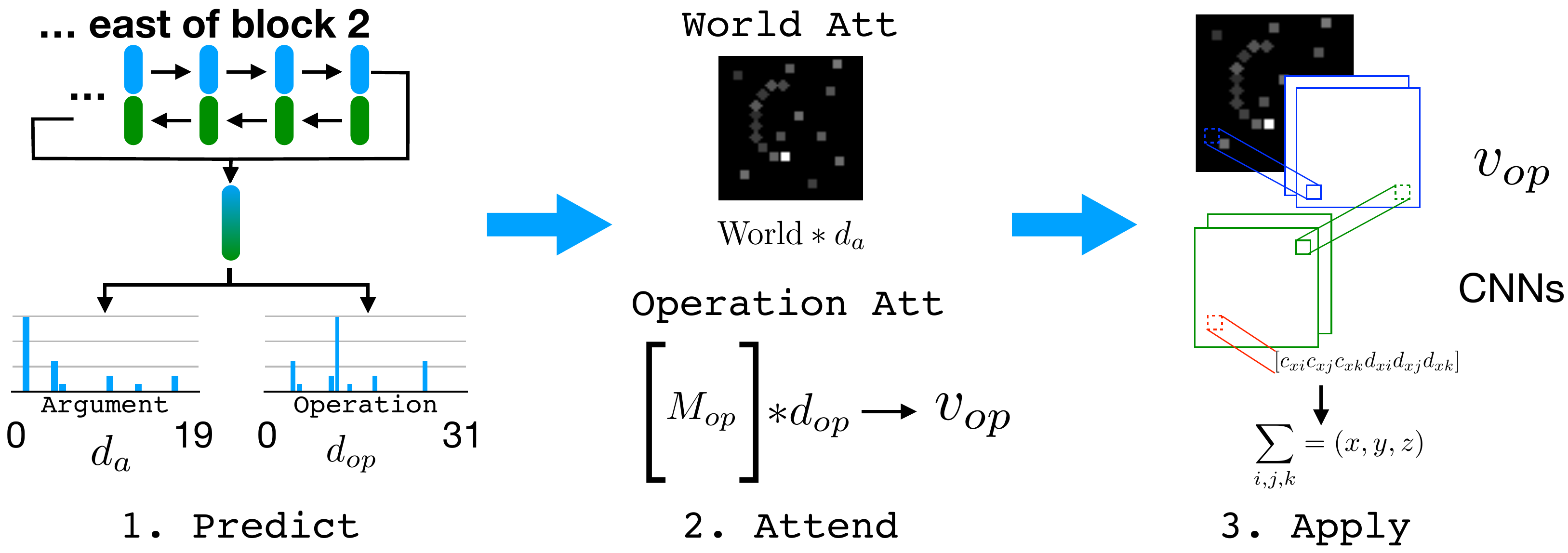}
\caption{Our target prediction model uses the sentence to produce distributions over operations and blocks (arguments).  The argument values illuminate regions of the world before the selected operation is applied.  This final representation is used to predict offsets in $(x, y, z, \theta)$ space. In practice, two bi-LSTMs were used and the final vector contains rotation information.}
\label{fig:model}
\end{figure*}

\subsection{Evaluation and Angles}
We follow the evaluation setup by prior work and evaluate 
by reporting the average distance ($L_2$ in block lengths) between where 
a block should be placed and the model's prediction.  This metric 
naturally extends to 3D.  
\begin{equation}
	L_{x,y,z}(p, g) = || p - g ||_2
\end{equation}

We also devise a metric for evaluating rotations.  In our released data,\footnote{\url{https://groundedlanguage.github.io/}} we
captured block orientations as quaternions.  This allows for a
complete and accurate re-rendering of the exact block orientations
produced by our annotators.  However, the most semantically
meaningful angle is the Eulerian rotation around the Y-axis.  We will
therefore evaluate error as the minimal angle between the ground truth and prediction in radians as:
\begin{equation}
	L_\theta(p, g) = atan2(sin(p - g), cos(p - g))
\end{equation}

\subsection{Example Phenomena}

In the following example, nine instructions (three per annotator) are provided 
for the proper placement of McDonald's.  We see a diverse set of concepts that
include counting, abstract notions like mirror or parallel,
geometric concepts like a square or row, and even constraints
specified by three different blocks.

\begin{tabular}{cc}
\textbf{$t_1$} & \textbf{$t_2$} \\
\includegraphics[width=0.35\linewidth]{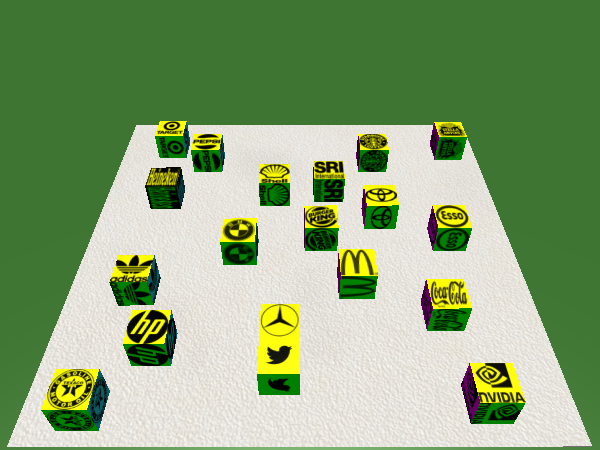} &
\includegraphics[width=0.35\linewidth]{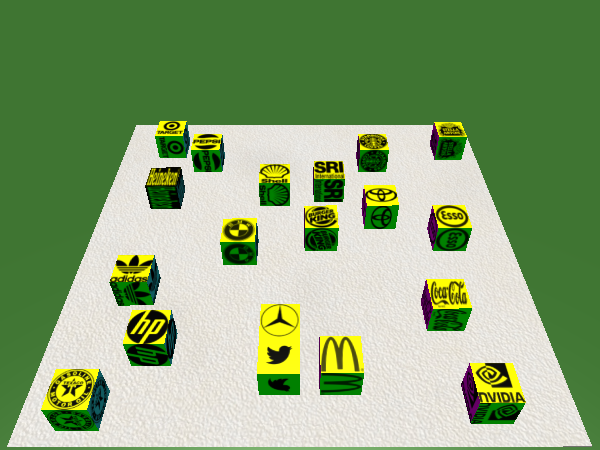} \\
\end{tabular}\\

\begin{footnotesize}
\begin{tabular}{l@{\hspace{6pt}}l@{}p{8cm}@{}}
\multicolumn{3}{l}{McDonalds...}\\
1 & ... & \textbf{mirrors} Twitter across the Y-axis.\\ 
2 & ... & to just over one space right of Twitter.\\ 
3 & ... & to the right of twitter with \textbf{1/2 block} in between\\
4 & ... & the right side of twitter with a \textbf{small space} in between\\ 
5 & ... & as the bottom right \textbf{square}, \textbf{parallel} with Twitter,\\
  &     & but a little further than touching.\\ 
6 & ... & so it's just to the right (not \textbf{touching}) the Twitter block.\\ 
7 & ... & \textbf{1/3-block's-length} to the right of the Twtter block.\\ 
8 & ... & will move down and right until it is in the same \textbf{row} \\
  &     & as twitter with a small space between them\\ 
9 & ... & move it downwards enough to line up with the Twitter \\
  &     & logo, then move it left to be closer to the Twitter logo, \\
  &     & but not touching. The McDonald's logo should appear \\
  &     & to be \textbf{inbetween the boundaries} of BurgerKing's \\
  &     & left edge and SRI's right edge. \\
\end{tabular}
\end{footnotesize}

Later in the same task, the agent will be asked to rotate a block and place
it between the two stacks. We present here just a few excerpts wherein the 
same action is described in five different ways.

\begin{tabular}{cc}
\centering
\textbf{$t_7$} & \textbf{$t_8$} \\
\includegraphics[width=0.35\linewidth]{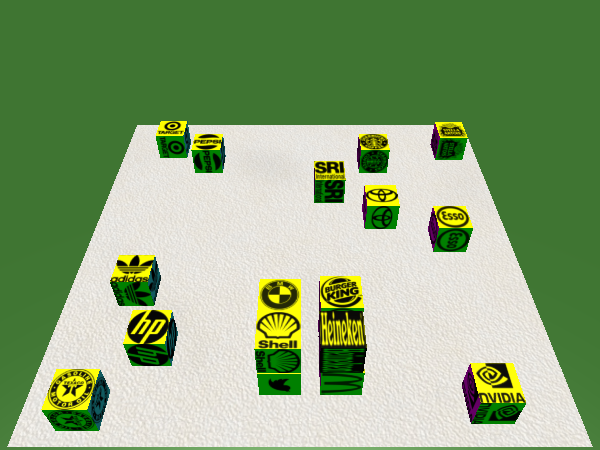} &
\includegraphics[width=0.35\linewidth]{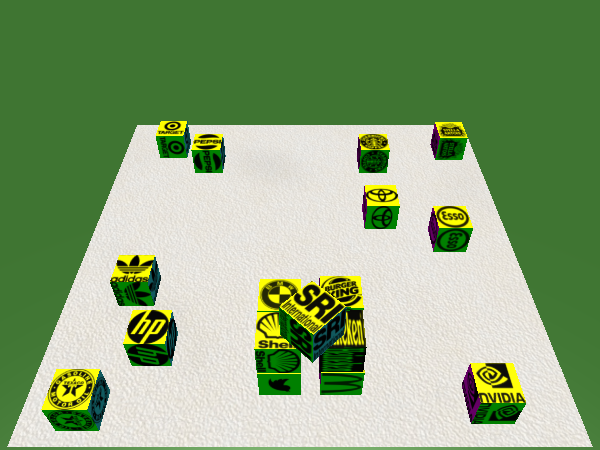} \\
\end{tabular}

\begin{footnotesize}
\centering
\begin{tabular}{l@{\hspace{6pt}}p{8cm}@{}}
1 & \textbf{Rotate} SRI to the \textbf{right} ... \\ 
2 & rotate it \textbf{45 degrees clockwise} ...\\ 
3 & \textbf{only half of one rotation} so its corners point \\
  & where its edges did ... \\ 
4 & \textbf{the logo faces the top right corner} of the screen...\\ 
5 & Spin SRI \textbf{slightly} to the right and then set it \\
  &  \textbf{in the middle of the 4 stacks}\\
\end{tabular}
\end{footnotesize}

To complete these instructions requires understanding angles, a new
set of verbs (\textit{rotate, spin, ...}), and references to the block's
previous orientation.  The final example, indicates that a spin is
necessary, but assumes the goal of having it balance between the two
stacks is sufficient information to choose the right angle.

The world knowledge and concepts necessary to complete this task are
well beyond the ability of any systems we are currently aware of or
expect to be built in the near future.  Our goal is to provide data
and an environment which more accurately reflects the complexity of
grounding language to actions.  Where previous work broadened
the community's understanding of the types of natural language people
use by recreating a blocks world with real human annotators, we felt
they did not go far enough in really covering the space of actions and
therefore language naturally present in even this constrained world.

\section{Model}
\label{sec:model}

In addition to our dataset, we propose an end-to-end trainable model that is both competitive in
performance and has an easily interpretable internal
representation. The model takes in a natural language instruction for
block manipulation and a 3D representation of the world as
input, and outputs where the chosen block should be moved. The model can be broken down into three primary components:
\begin{enumerate}
\item Language Encoding for Block and Operation prediction
\item Applying a spatial operation
\item Predicting a coordinate in space.
\end{enumerate}

Our overall model architecture is shown in Figure \ref{fig:model}.
By keeping the model modular we can both control the bottlenecks that
learning must use for representation and provide ourselves post hoc access
to interpretable scene and action representations (explored further in interpretability section).  Without these, the model allows sentences 
and operations to be represented by arbitrary N-dimensional vectors.

\subsection{Language Encoder}
As is common, we use bidirectional LSTMs~\cite{hochreiter1997long,schuster1997bidirectional} to encode the input sentence.
We use two LSTMs: one for predicting blocks to attend to, one for 
choosing the operations to apply. Both LSTMs share a vocabulary embedding matrix, but
have no other means of communication.  We experimented with using a single
LSTM as well as conditioning one on the other, but found it degraded performance.

Once we have produced a representation for arguments $h_a$ and operations $h_o$, 
we multiply each by their own feed-forward layers, then softmax to
produce a distribution over 20 blocks and 32 operations for $d_a$ and $d_{op}$, respectively.

\begin{equation}
\begin{aligned}
  d_a &= \mathrm{softmax(}W_a h_a + b_a)\\
  d_{op} &= \mathrm{softmax(}W_o h_o + b_o)
\end{aligned}
\end{equation}

\paragraph{Argument Softmax}
The first output of our model is an attention over the block IDs.  
The input world is represented by a
3D tensor of IDs.\footnote{In principle, we could work over an RGB rendering of the
world, but doing so would add layers of vision complexity that
do not help address the dominant language understanding problems.}
We can convert this to a one-hot representation and
multiply it by the distribution to get an attention per ``pixel'' (hereby
referred to as argument attention map) equal to the model's
confidence.  In practice we found that the model was better able to
learn when the attention map was multiplied by 10. This may be due to parameter 
initialization. Additionally, we do not allow the model to attend to 
background so it is masked out (result in Figure \ref{fig:model}) We use the 
operator * to represent the inner product.

\begin{equation}
    \begin{aligned}
    A_{i,j,k}& = 10 ( \mathrm{\texttt{one\_hot}}(\mathrm{World}_{i,j,k}) * d_a)\\
    A_{i,j,k}& = A_{i,j,k}*\mathrm{\texttt{bg\_mask}}_{i,j,k}
    \end{aligned}
\end{equation}

\paragraph{Operation Softmax}
The second distribution we predict is over functions for spatial relations.  Here
the model needs to choose how far and in what directions to go from
the blocks it has chosen to focus on.  Unfortunately, there is no a
priori set of such functions as we have specifically chosen not to try and
pretrain/bias the model in this capacity, so the model must perform a type of
clustering where it simultaneously chooses a weighted sum of functions
and trains their values.

As noted previously, for the sake of interpretability, we force the
encoding for operations ($d_{op}$) to be a latent softmax distribution
over 32 logits. The final operation vector that is passed along to the
convolutional model is computed as:
\begin{equation}
  v_{op} = M_{op}d_{op}
\end{equation}

Here, $M_{op}$ is a set of 32 basis vectors. The output vector
$v_{op}$ is a weighted average across all 32 basis vectors, using
$d_{op}$ to weight each individual basis. The goal of this formulation
is such that each of the 32 basis vectors will be independently
interpretable by replacing $d_{op}$ with a 1-hot vector, allowing us
to see what type of spatial operation each vector represents. 
The choice of 32 basis vectors was an empirical
one.  We only experimented with powers of two, but it is quite likely
a more optimal value exists.

\subsection{Predicting a location}
The second half of our pipeline features a convolutional model that
combines the encoded operation and argument blocks with the world
representation to determine the final location of the block-to-move.

Given the aforementioned argument attention map (tensor $A$ of size
$B \times D \times H \times W \times 1$, our model starts by applying
the operation vector $v_{op}$ at every location of the map, weighted
by each location's attention score. This creates a world
representation of size $B\times D \times H \times W \times
|v_{op}|$. We then pass this world through two convolutional layers
using \texttt{tanh} or \texttt{relu} nonlinearities.

In order to predict the final location for the block-to-move, we apply
a final $1\times 1\times 1$ convolutional layer to predict offsets and their respective
confidences for each location relative to a coordinate grid (8 values total). The
coordinate grid is a constant 3D tensor generated by uniformly
sampling points across each coordinate axis to achieve the desired
resolution. Given the coordinate grid, the goal of the learned
convolutional model is to, at every sampled point, predict offsets for
$x$, $y$, $z$, $\theta$, as well as a confidence for each predicted
offset. This formulation was similarly used for keypoint localization in~\cite{singh2016learning}. Let $g_x(i,j,k)$ be the $x$ coordinates for all sampled grid
points at grid location $(i,j,k)$ and let $d_x(i,j,k)$ and
$c_x(i,j,k)$ be the respective offsets and confidences,
then the final predicted $\hat{x}$ coordinate for the block-to-move
is computed as: 
\begin{equation}
\hat{x}=\sum_{i,j,k}c_x(i,j,k) * (g_x(i,j,k) + d_x(i,j,k))
\end{equation}

Here, confidences $c_x(i,j,k)$ are softmax normalized across all grid
points. Predictions for $\hat{y}$, $\hat{z}$ are computed similarly.  We compute $\hat{\theta}$ without a coordinate grid such that: $\hat{\theta}=\sum_{i,j,k}c_{\theta}(i,j,k)d_{\theta}(i,j,k)$.

\subsection{Implementation Details}
Our model is trained end-to-end using Adam~\cite{kingma2014adam} with
a batch size of 32.
The convolutional aspect of the model has
3 layers and operates on a world representation of
dimensions 32 $\times$ 4 $\times$ 64 $\times$ 64 $\times$ 32 (batch, depth, height,
width, channels). The first
convolutional layer uses a filter of size 4 $\times$ 5 $\times$ 5 and
the second of size 4 $\times$ 3 $\times$ 3, each followed by
a \texttt{tanh} nonlinearity for the 3D model\footnote{We did not perform a grid search for parameters, but we did find the 2D model performed better when a \texttt{relu} was used and Batch-Normalization \cite{ioffe15}.  Finally, the depth values and kernel were set to 1 when training exclusively in 2D.}.
Both layers
output a tensor with the same dimensions as the input world. The final
predicton layer is a 1 $\times$ 1 $\times$ 1 filter that projects the
32 dimensional vector at each location down to 8 values as detailed in
the previous section. We further include an entropy term to encourage peakier distributions in the argument and operation softmaxes.

\begin{figure}
\centering
\includegraphics[width=0.6\linewidth]{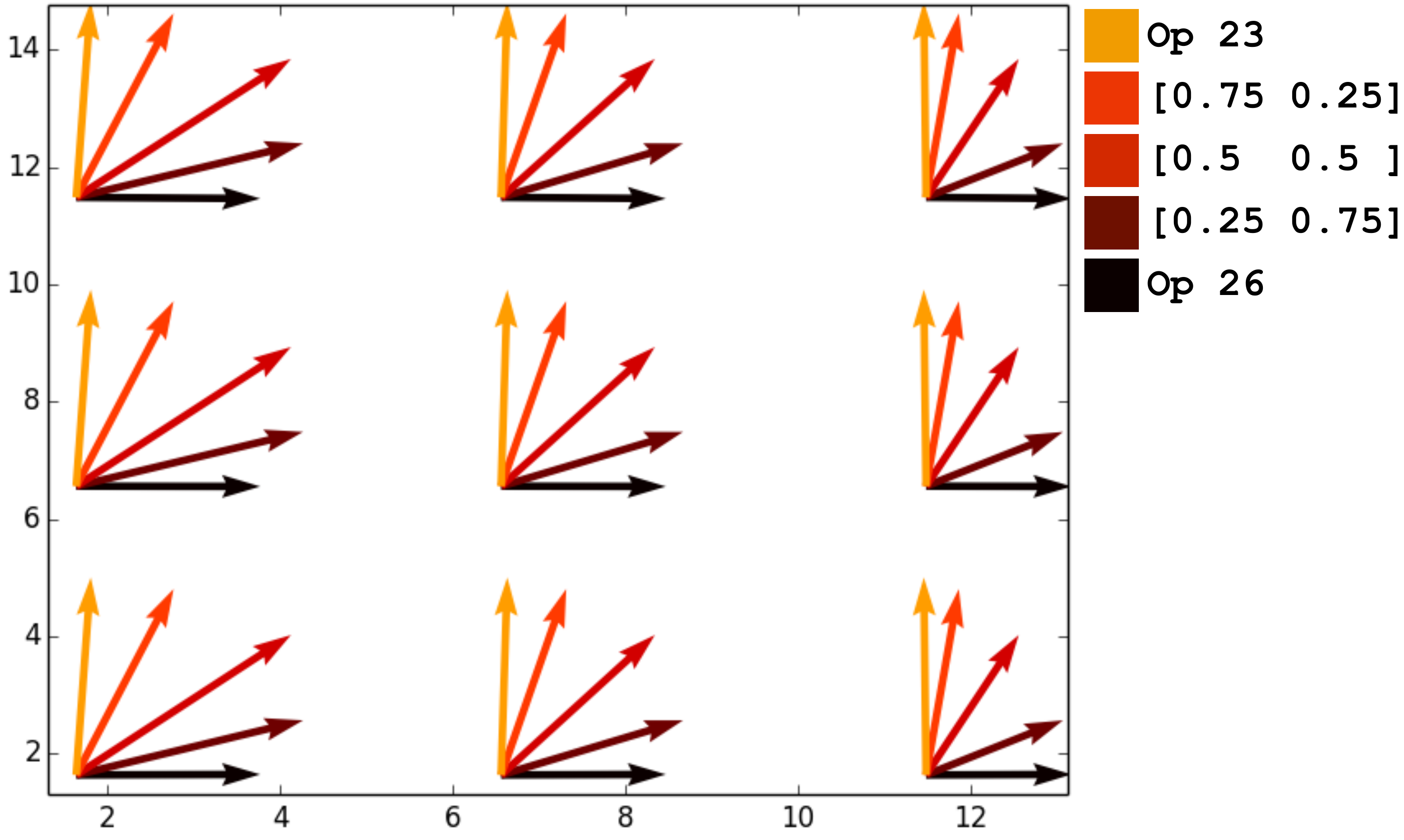}
\caption{Interpolations of operations 23 (north) and 26 (east) being applied at nine locations around the world.}
\label{fig:op_interp}
\end{figure}

\begin{figure*}
\centering
{\scriptsize \texttt{~0}} \includegraphics[width=0.22\textwidth]{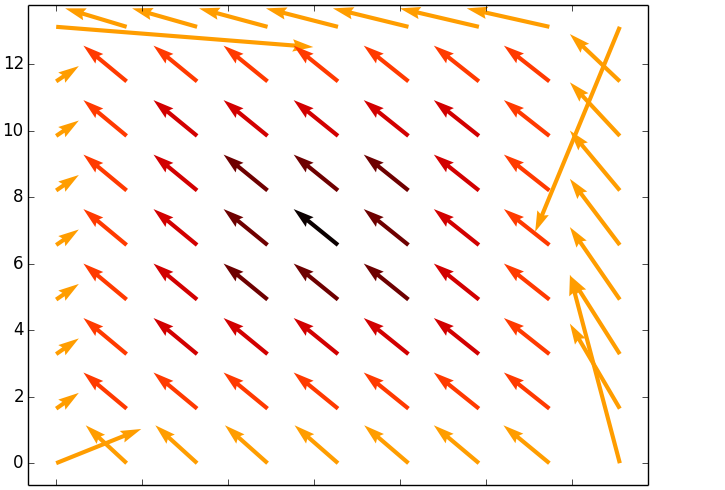}
{\scriptsize \texttt{~1}} \includegraphics[width=0.22\textwidth]{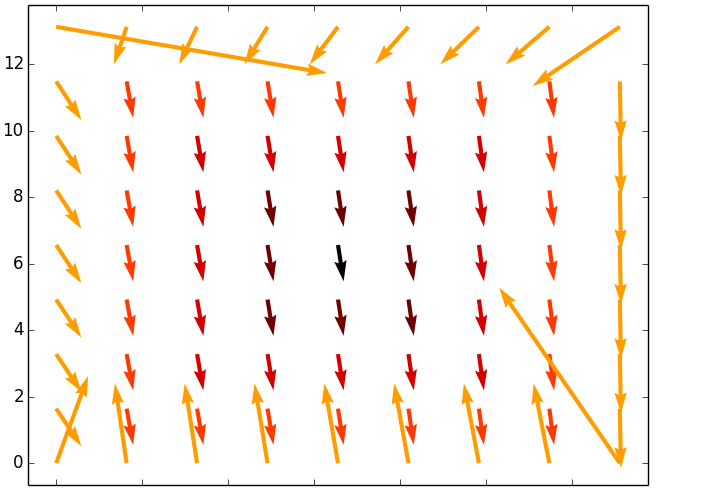}
{\scriptsize \texttt{~2}} \includegraphics[width=0.22\textwidth]{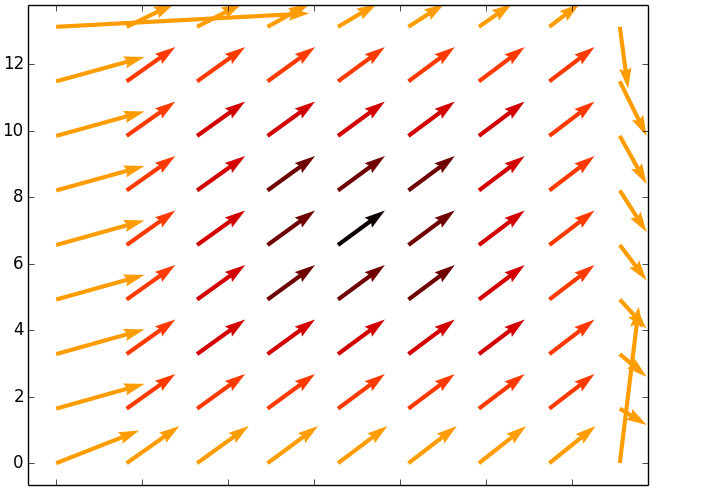}
{\scriptsize \texttt{~3}} \includegraphics[width=0.22\textwidth]{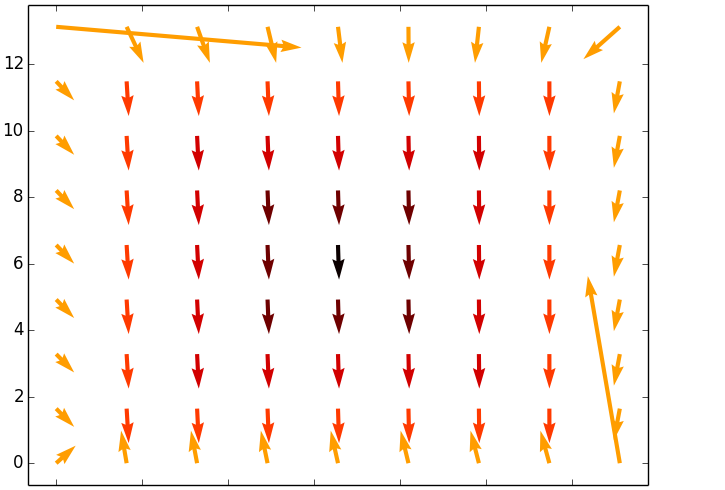}

{\scriptsize \texttt{~4}} \includegraphics[width=0.22\textwidth]{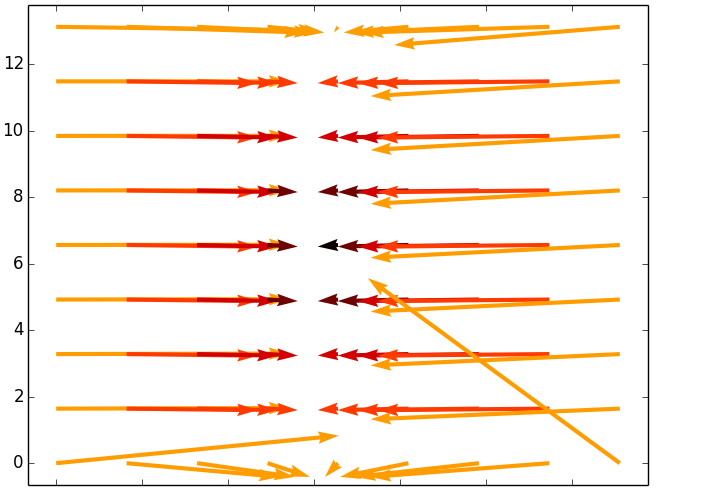}
{\scriptsize \texttt{~5}} \includegraphics[width=0.22\textwidth]{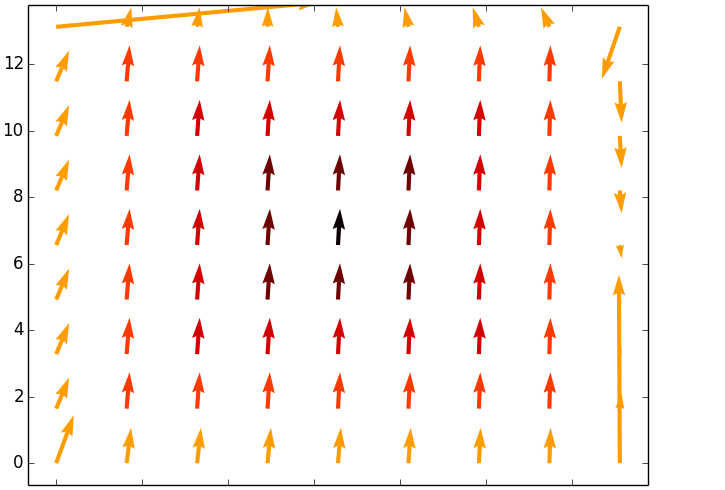}
{\scriptsize \texttt{~6}} \includegraphics[width=0.22\textwidth]{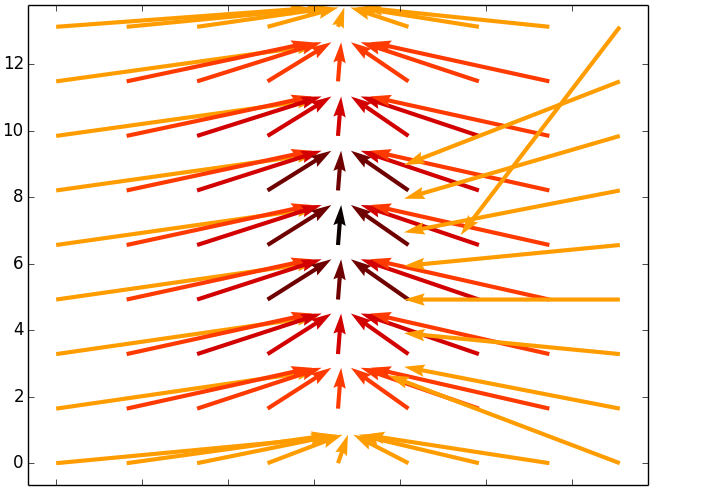}
{\scriptsize \texttt{~7}} \includegraphics[width=0.22\textwidth]{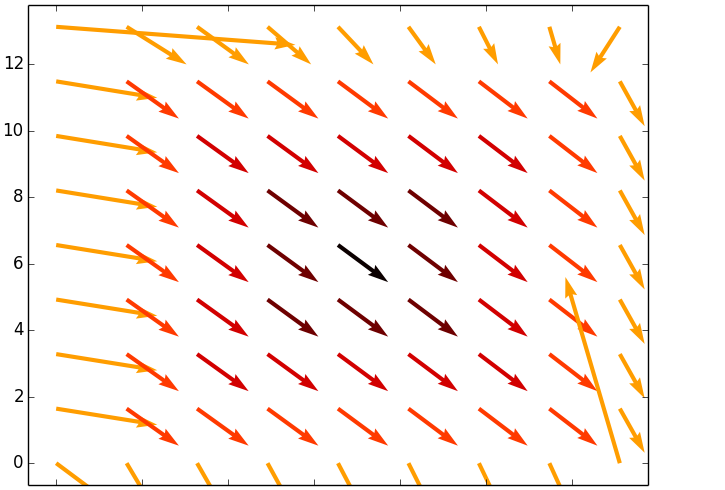}

{\scriptsize \texttt{~8}} \includegraphics[width=0.22\textwidth]{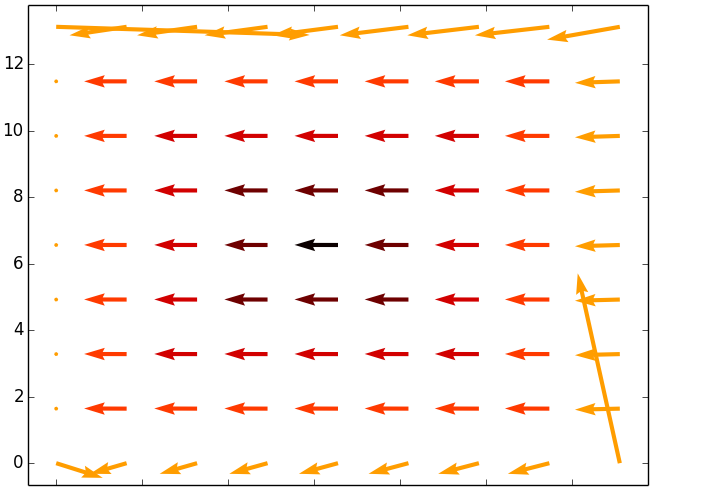}
{\scriptsize \texttt{~9}} \includegraphics[width=0.22\textwidth]{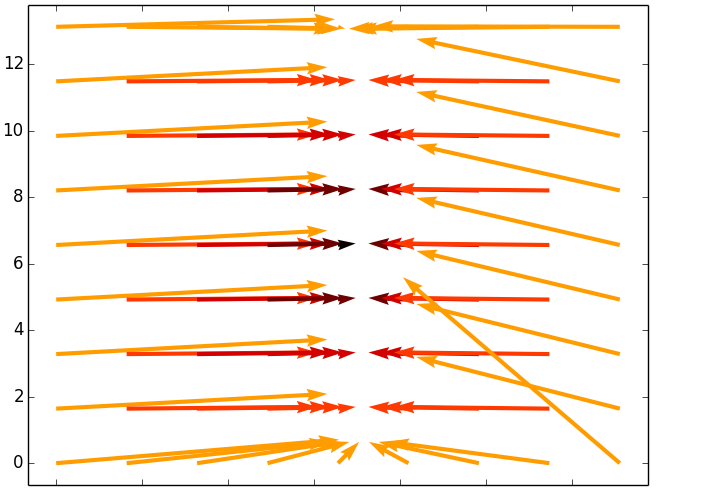}
{\scriptsize \texttt{10}} \includegraphics[width=0.22\textwidth]{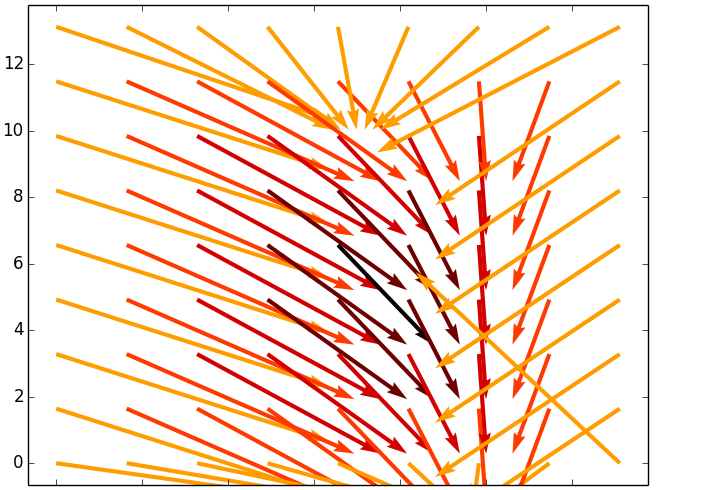}
{\scriptsize \texttt{11}} \includegraphics[width=0.22\textwidth]{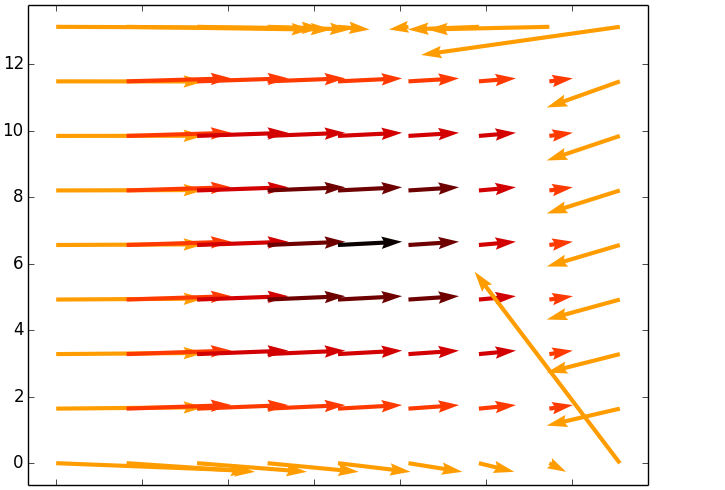}

{\scriptsize \texttt{12}} \includegraphics[width=0.22\textwidth]{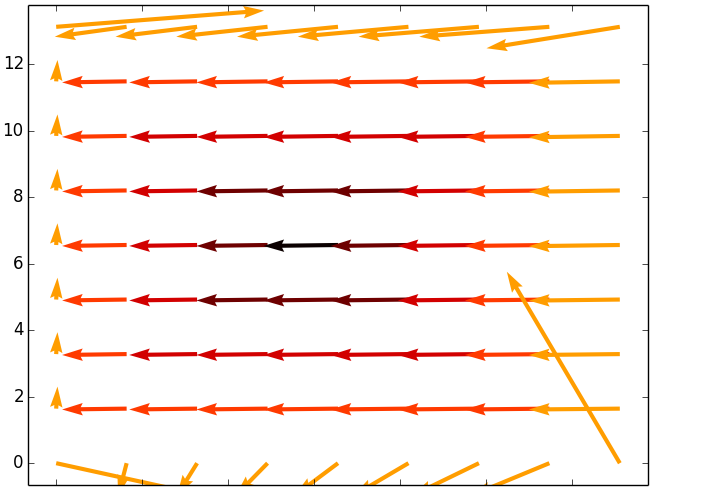}
{\scriptsize \texttt{13}} \includegraphics[width=0.22\textwidth]{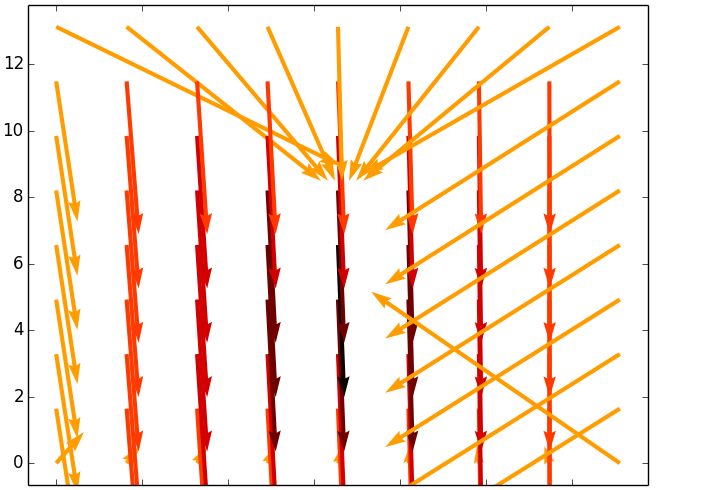}
{\scriptsize \texttt{14}} \includegraphics[width=0.22\textwidth]{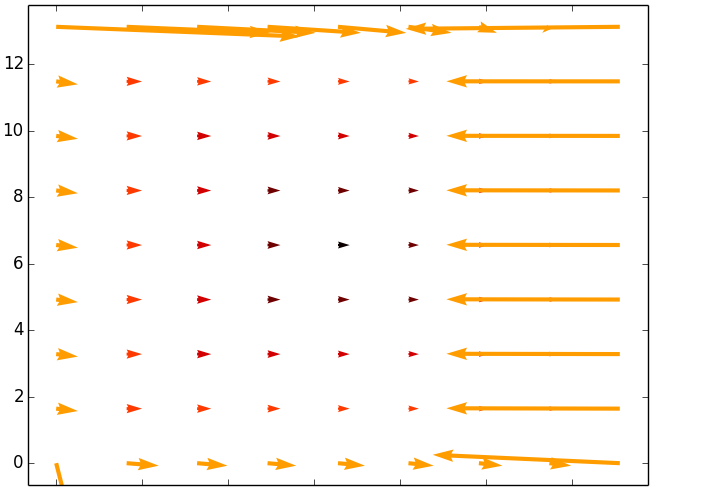}
{\scriptsize \texttt{15}} \includegraphics[width=0.22\textwidth]{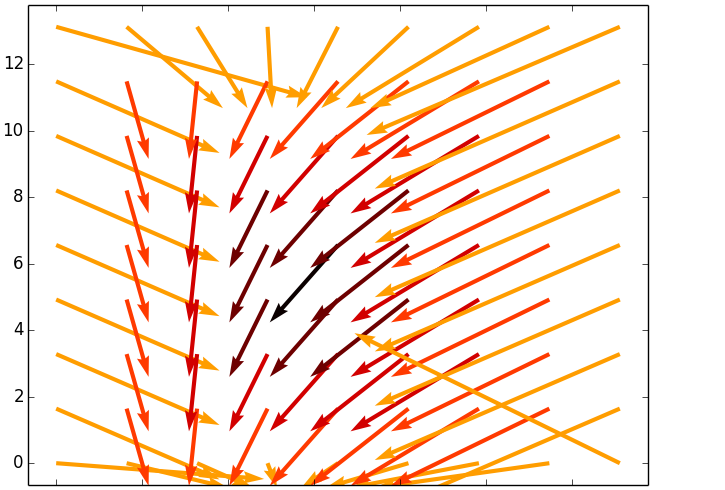}

{\scriptsize \texttt{16}} \includegraphics[width=0.22\textwidth]{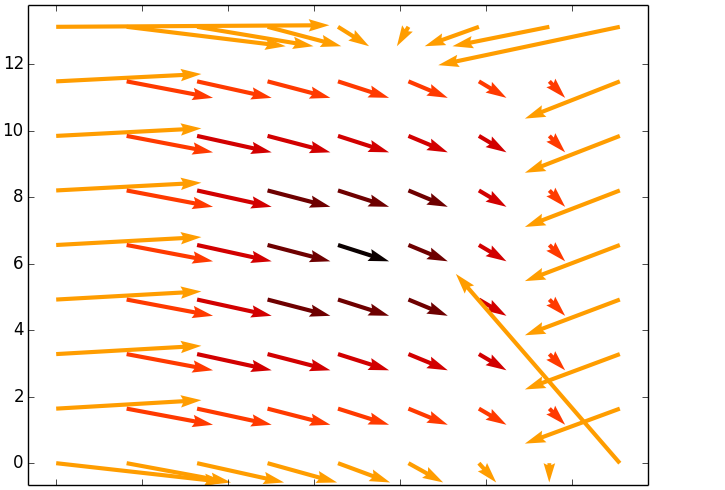}
{\scriptsize \texttt{17}} \includegraphics[width=0.22\textwidth]{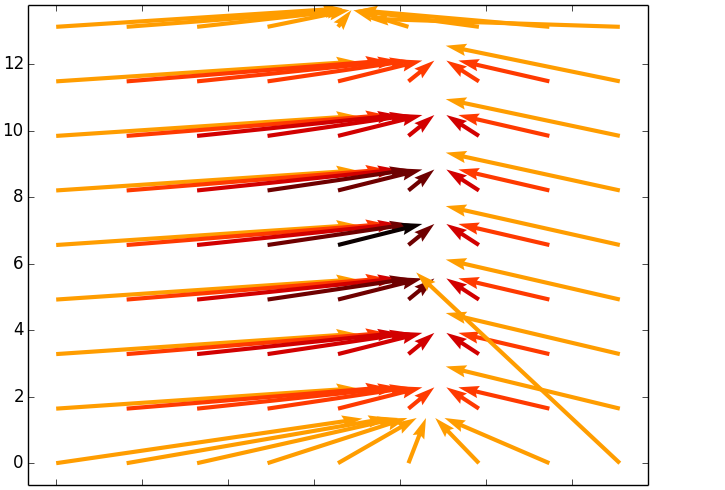}
{\scriptsize \texttt{18}} \includegraphics[width=0.22\textwidth]{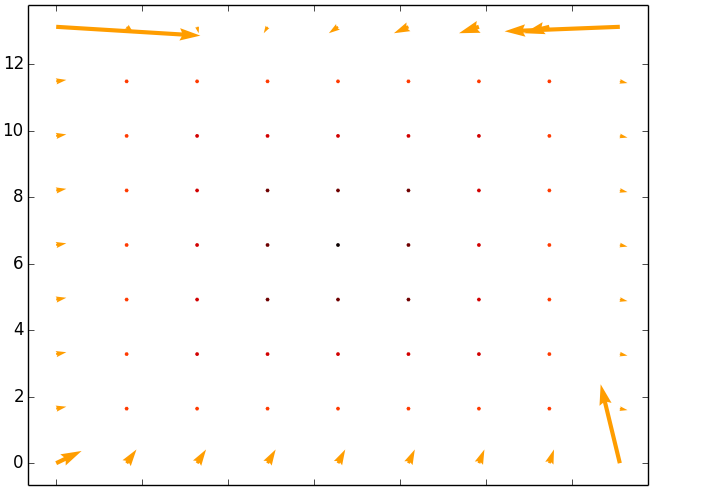}
{\scriptsize \texttt{19}} \includegraphics[width=0.22\textwidth]{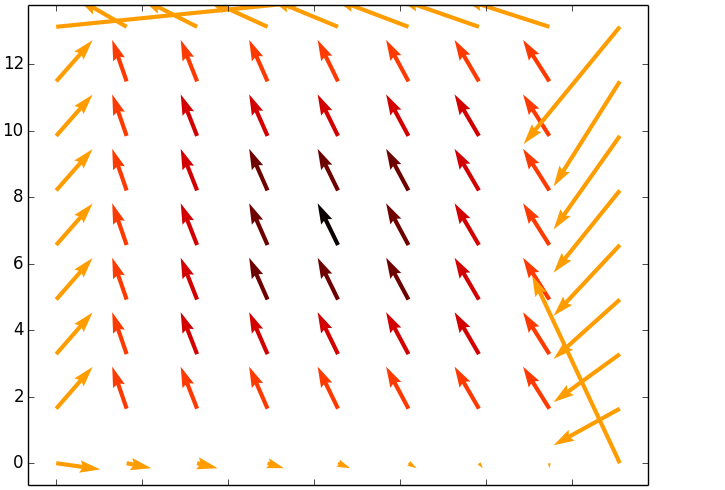}

{\scriptsize \texttt{20}} \includegraphics[width=0.22\textwidth]{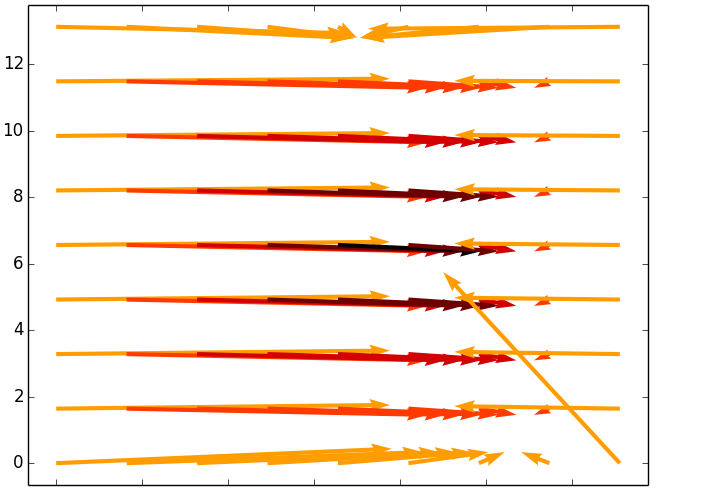}
{\scriptsize \texttt{21}} \includegraphics[width=0.22\textwidth]{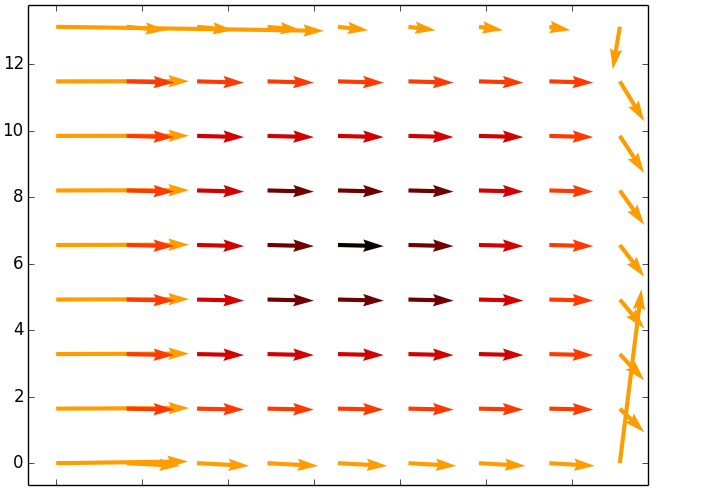}
{\scriptsize \texttt{22}} \includegraphics[width=0.22\textwidth]{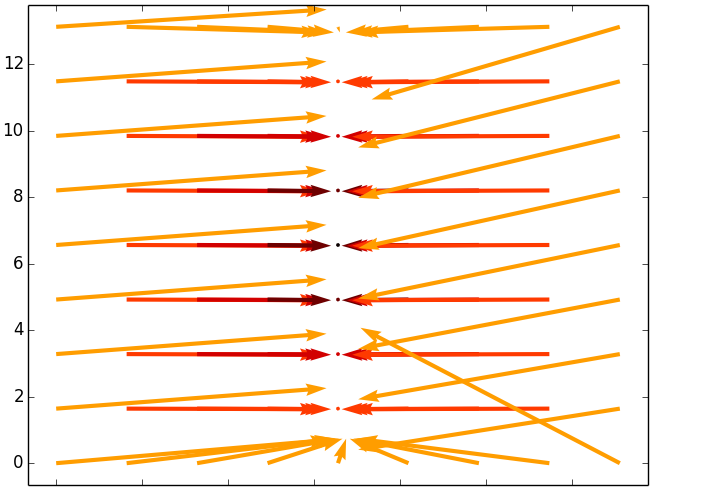}
{\scriptsize \texttt{23}} \includegraphics[width=0.22\textwidth]{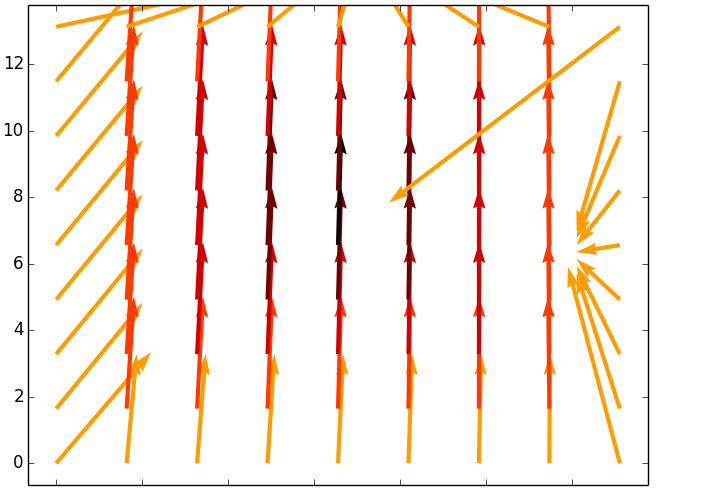}

{\scriptsize \texttt{24}} \includegraphics[width=0.22\textwidth]{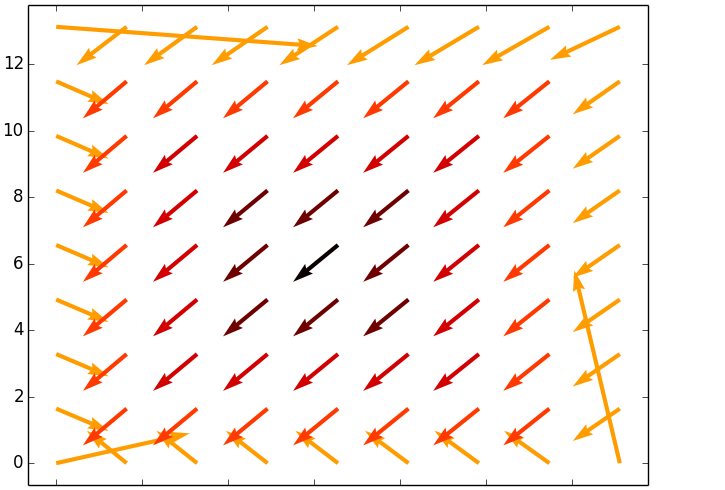}
{\scriptsize \texttt{25}} \includegraphics[width=0.22\textwidth]{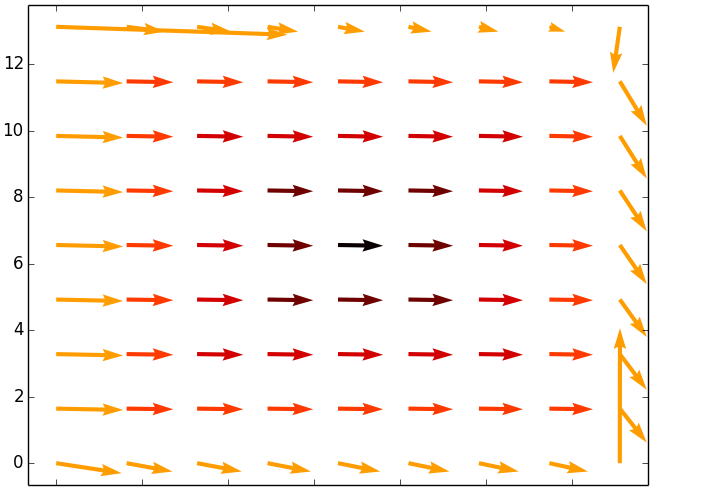}
{\scriptsize \texttt{26}} \includegraphics[width=0.22\textwidth]{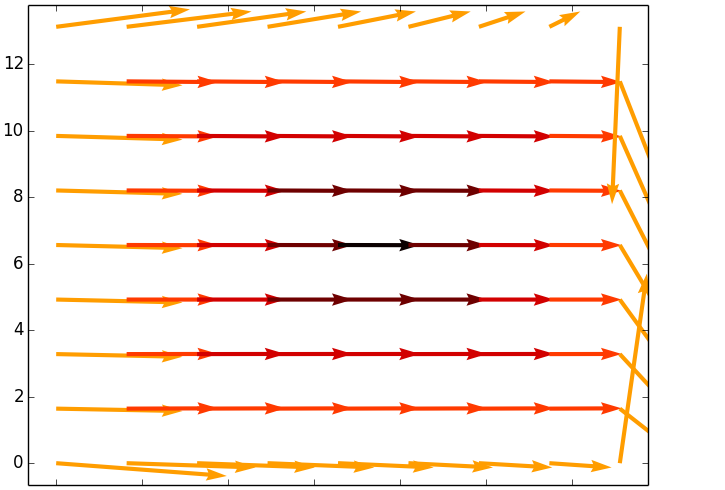}
{\scriptsize \texttt{27}} \includegraphics[width=0.22\textwidth]{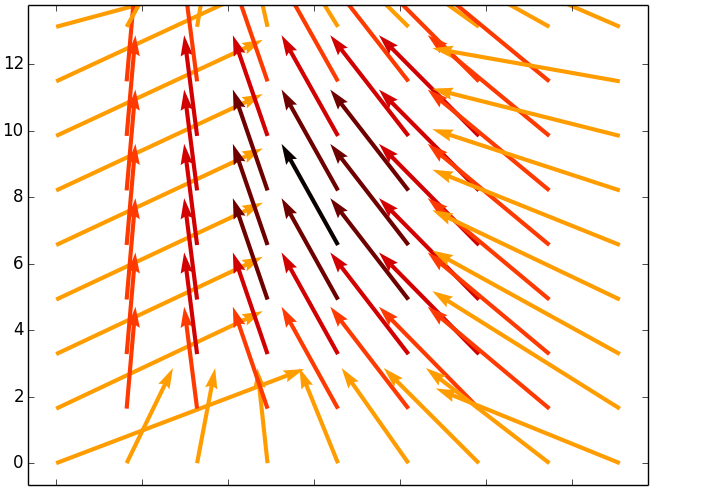}

{\scriptsize \texttt{28}} \includegraphics[width=0.22\textwidth]{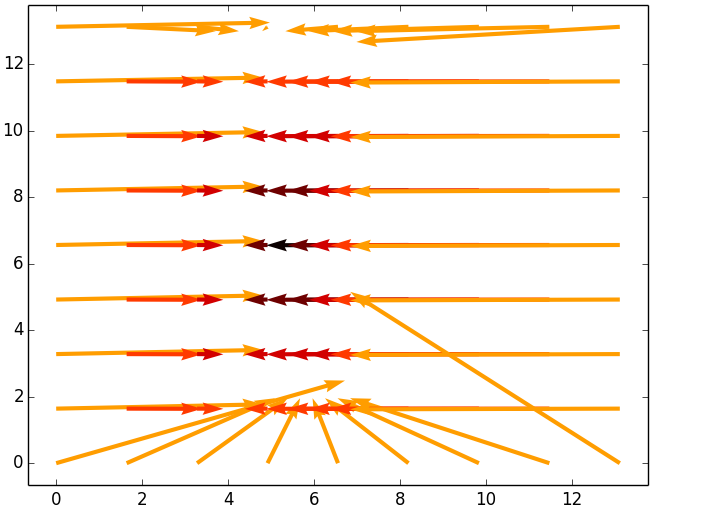}
{\scriptsize \texttt{29}} \includegraphics[width=0.22\textwidth]{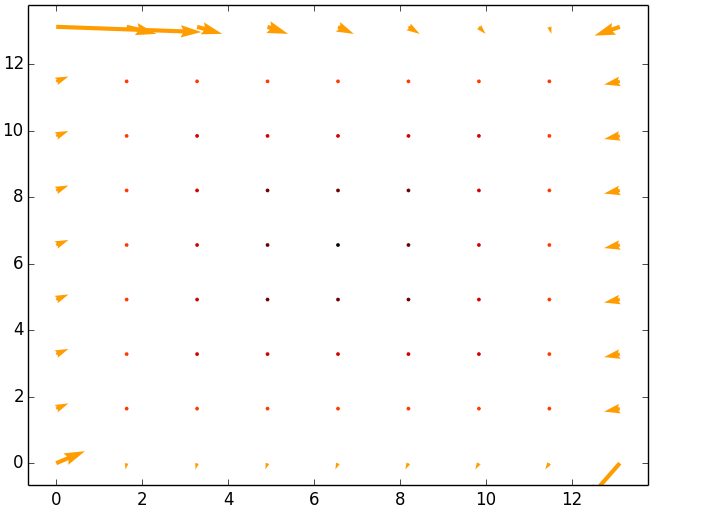}
{\scriptsize \texttt{30}} \includegraphics[width=0.22\textwidth]{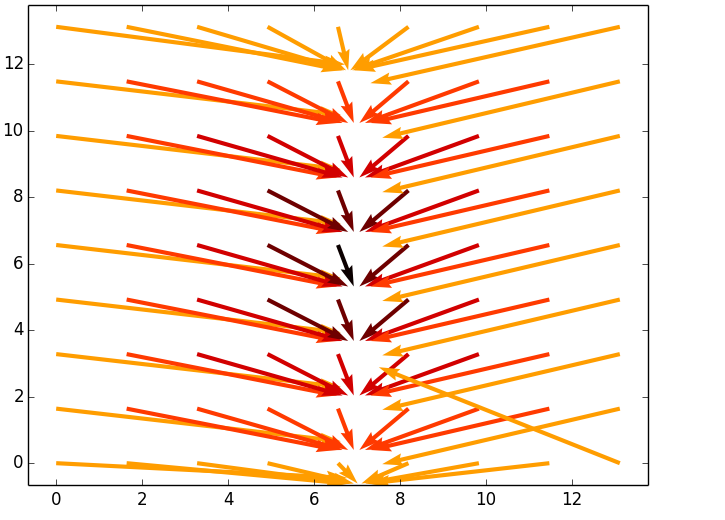}
{\scriptsize \texttt{31}} \includegraphics[width=0.22\textwidth]{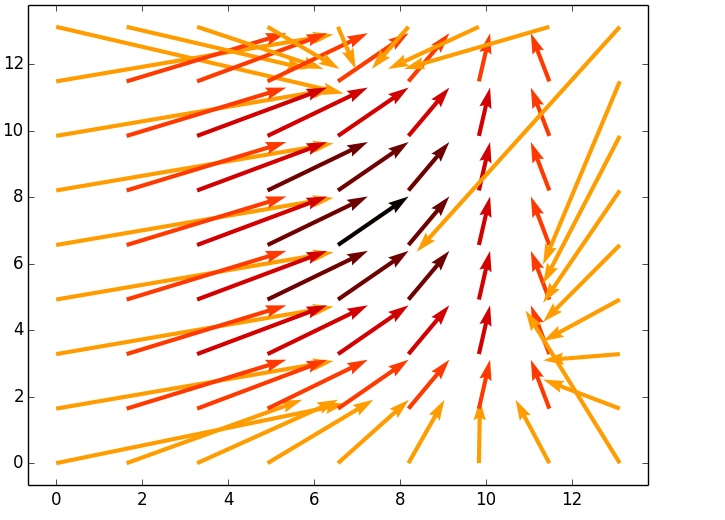}

\caption{A 2D projected visualization of the operations our model discovered.  Common operations are described in Table \ref{tab:attention}.  Short arrows are mostly in 3D, and nearly all operations exhibit different behaviors depending on location in the world.}
\label{fig:operations_grid}
\end{figure*}

\section{Interpretability and Visualizing the Model}
One of the features of our model is its interpretability, which we
ensured by placing information bottlenecks within the architecture. By
designing the language-to-operation encoding process as predicting a
probability distribution over a set of learned basis vectors, we can interpret
each vector as a separate operation and visualize the behaviors of each operation 
vector individually. 

\paragraph{Visualizing Operations}
We generated Figure \ref{fig:operations_grid} by placing a single block
in the world and moving it around a 9 by 9 grid and passing
a 1-hot operation choice vector to our model. We then plot a vector from
the block's center to the predicted target location.  We see many simple and expected
relationships (left, right, ...), but importantly we see the operations are
location specific functions, not simply offsets.  Operations on the edges of
the world are more fine-grained and many move directly to a region
of the world (e.g. 9 = ``center''), not simply an offset.  It is also possible that 
some of the more dramatic edge vectors may serve as a failsafe mechanism for misinterpreted
instructions.  In particular, nearly all of the operations when applied in the bottom right 
corner redirect to the center of the board rather than off of it.

Additionally, while shown here in 2D, all of our predictions are actually in 3D and contain rotation predictions.
In Figure \ref{fig:operations_grid} the operations denoting directly on top are the figures with the shortest arrows (e.g. Operation 14).

\label{sec:interpretability}
\begin{table}
\centering
\begin{footnotesize}
\begin{tabular}{r|lll@{}}
OP&\multicolumn{3}{l}{Descriptions learned from our data}\\
\hline
0& directly northwest & left and above & upper left \\
\rowcolor{lgray}1& directly below & directly under & \multicolumn{1}{l}{south} \\
\multirow{2}{*}{2} & \multicolumn{3}{l}{bottom left corner should be almost touching upper right}\\
& northeast & &\\
\rowcolor{lgray}3& south of & \multicolumn{2}{l}{below}\\
4& next to\\
\rowcolor{lgray}& bottom side should touch & directly above & \multicolumn{1}{l}{north of} \\
\multirow{2}{*}{7}& below and to the right of & top left corner & \\
& \multicolumn{3}{l}{is almost touching the lower right corner of}\\
\rowcolor{lgray}8& left of & \multicolumn{2}{l}{directly west of}\\
\multirow{3}{*}{11}& \multicolumn{3}{l}{shifted between block hp and block nvidia}\\
& \multicolumn{3}{l}{its center with the line between the nvidia and mercedes }\\
\rowcolor{lgray}14& goes on top of  & \multicolumn{2}{l}{stacked on top} \\
18& \multicolumn{3}{l}{place it on top of}\\
\rowcolor{lgray}19& \multicolumn{3}{l}{directly under it}\\
21& to the right of& \multicolumn{2}{l}{directly on the right of}\\
\rowcolor{lgray}23& 5 block lengths above & \multicolumn{2}{l}{three rows above} \\
24& southwest of& \multicolumn{2}{l}{diagonally below}\\
\rowcolor{lgray}25& east of& \multicolumn{2}{l}{to the right of}\\
& \multicolumn{3}{l}{on the east side of the nvidia cube}\\
\multirow{-2}{*}{26}& \multicolumn{3}{l}{so that there is no space in between them}\\
\rowcolor{lgray}& \multicolumn{3}{l}{three spaces to the left of}  \\
\rowcolor{lgray}\multirow{-2}{*}{28}& \multicolumn{3}{l}{to the left with two intervening empty block spaces}\\
& \multicolumn{3}{l}{bottom face touches bmw's top face}\\
\multirow{-2}{*}{29}& cover shell with heineken & \multicolumn{2}{l}{top of it}\\
\end{tabular}
\end{footnotesize}
\caption{Utterances with low entropy for Op predictions were mapped to their corresponding argmax dimension. We extracted relevant phrase here for common dimensions.}
\label{tab:attention}
\end{table}

\paragraph{Interpolating Operations}
The 1-hot operations can be treated like API calls
where several can be called at the same time and interpolated. 
Figure~\ref{fig:op_interp} shows the predicted offsets when interpolating
operations 23 (north) and 26 (east).  There are two important takeaways
from this.  First, we see that when combined, we can sweep out angles in the
first quadrant to reference them all.  Second, we see that magnitudes and 
angles change as we move closer to the edges of the world.  This result
is intuitive and desired.  Specifically, a location like ``to the right" has
a variable interpretation depending on how much space exists in the world,
and the model is trying to make sure not to push a block off the table.
In practice, our analysis found very few clear cases of the model using this power.
More commonly, mass would be split between two very similar operations or the
sentence was a compound construction (\textit{left of X and above Y}).  We did find 
that operation 11 correlated with the description between but it is difficult to 
divine why from the grid. An important
extension for future work will be to construct a model which can apply multiple
operations to several distinct arguments.

\paragraph{Linguistic Paraphrase}
Using the validation data, we clustered sentences by their predicted operation
vectors.  To pick out phrases we only look at sentences with very low entropy
distributions (highly confident) and we present our findings in Table \ref{tab:attention}.
We see that specifications range from short one word indicators (e.g. \textit{below}) to 
nearly complete sentences (\textit{on the east side of the nvidia cube so that there is no space in between them}).  This also touches on the fact that several operations have the same direction but different magnitudes.  Specifically, operation 23 means far above, not directly, and we see this in the visualized grid as well.

\begin{table}
\centering
\begin{tabular}{ll@{\hspace{6pt}}l@{\hspace{6pt}}ll@{\hspace{6pt}}l}
\toprule
         & Source & \multicolumn{4}{c}{Target} \\
         &       & \multicolumn{2}{@{}c}{Gold Source} & \multicolumn{2}{c@{}}{End-to-End} \\
         & Acc. & Mean & Med &  Mean & Med  \\
\midrule
Bisk 16     & 98   & --  & --   &  0.98 & 0.0  \\ 
Pi\v{s}l 17 & 98.5 & --  & --   &  0.72 & --    \\
Ours        & 97.5 & 0.7 & 0.14 &  0.80 & 0.14  \\ 
\midrule
v2         & 91.3 & 1.2 & 0.85 &  1.15  & 0.88  \\ 
\midrule
v1 + v2    & 95.9 & 1.0 & 0.50 &  1.10  & 0.51  \\
v1 + v2 $\rightarrow$ v1 & 98.1 & 0.8 & 0.15 &  0.84 & 0.15 \\
v1 + v2 $\rightarrow$ v2 & 93.1 & 1.2 & 0.88 &  1.35 & 0.91 \\
\bottomrule
\end{tabular}
\caption{A comparison of our interpretable model with previous results (top) in addition to our performance on our new corpus (v2).  Finally, we show how training jointly on both corpora has only a very moderate effect on performance, indicating the complementarity of the data. Target values are error measurements in block-lengths (lower is better).}
\label{tab:results}
\end{table}

\section{Results}
\label{sec:results}
In Table \ref{tab:results}, we compare our model against existing work, and evaluate on both the original Blocks data (v1) and our new corpus (v2).  While our primary goal was the creation of an interpretable model and the introduction of new spatial and linguistic phenomena, it is important to see that our model also performs well.  We note three important results: 

First, we see that our model outperforms the original model of Bisk 16, and is only slightly weaker than Pi\v{s}l 17.  Our technique does outperform theirs when given the correct source block, so it is possible that we can match their performance with tuning.

Second, our results indicate that the new data (v2) is harder than v1, both in terms of isolating the correct block to move (91 vs 98\% accuracy) and average error (1.15 vs 0.80) on the End-to-End setting. Further, a model trained on the union of our corpora improved in source prediction on both the v1 and v2 test sets, but target location performance was either unaffected or slightly deteriorated.  This indicates to us that the new dataset is in fact complementary and adds new constructions.

Finally, our model has an average error of 0.058 radians (three degrees).  In validation, 46\% of predictions require a rotation. 1,374 of 1491 predictions are within 2 degrees of the correct orientation.  The remainder have dramatically larger errors (36 at 30$^{\circ}$, 81 at 45$^{\circ}$).  This means that the model is learning to interpret the scene and utterance correctly in the vast majority of cases.

\section{Error Analysis}
\label{sec:analysis}
Several of our model's worst performing examples are included in Table \ref{tab:analysis}.  The model's error is presented alongside the goal configuration and misunderstood instruction.
\begin{table}
\begin{footnotesize}
\begin{tabular}{@{}m{1cm}@{\hspace{3pt}}lb{5.2cm}@{}}
\toprule
Error & Goal & Instruction \\
\midrule
4.8 &\includegraphics[width=0.2\linewidth]{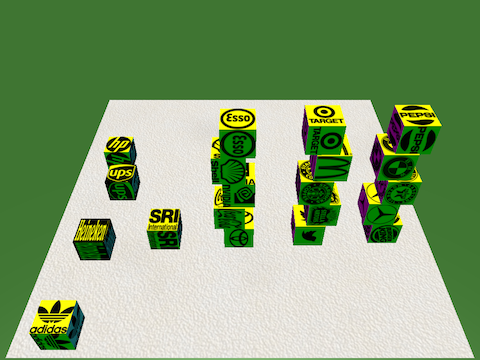} & use sri as
the base of a fourth tower to the left and equidistant with the other
tower\\ 5.2
& \includegraphics[width=0.2\linewidth]{images/spaceship_8} & spin sri
slightly to the right and then set it in the middle of the 4 stacks\\
6.4 & \includegraphics[width=0.2\linewidth]{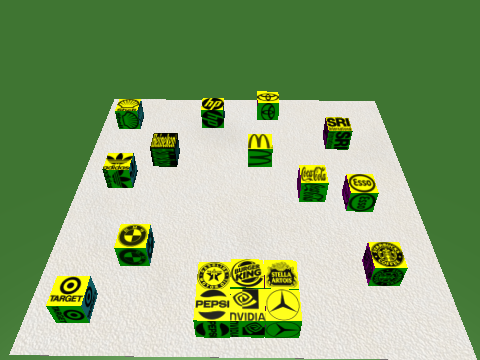} & in the
emerging 3x3 grid place texaco in the middle left\\
\bottomrule
\end{tabular}
\end{footnotesize}
\caption{Several of our worst performing results.  Errors are in block lengths, the images are the goal configuration, and the instructions have been lowercased and tokenized.}
\label{tab:analysis}
\end{table}

The first example specifies the goal location using an abstract concept (tower) and the offset (equidistant) implies recognition of a larger pattern.  The second example specifies the goal location in terms of ``the 4 stacks'', again without naming any of them and in 3D.  Finally, the third demonstrates a particularly nice phenomenon in human language where a plan is specified, the speaker provides categorizing information to enable its recognition, and then can use this newly defined concept as a referent.  No models to our knowledge have the ability to dynamically form new concepts in this manner.

\begin{table}
\centering
\begin{tabular}{@{\hspace{20pt}}cc}
\includegraphics[width=0.35\linewidth]{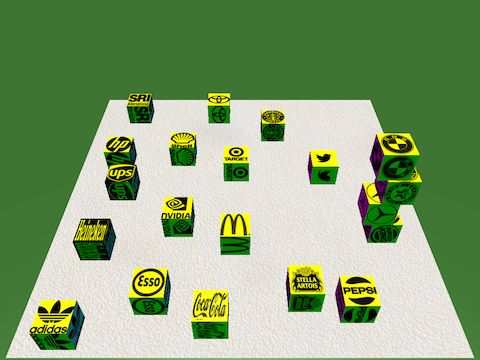} & \includegraphics[width=0.35\linewidth]{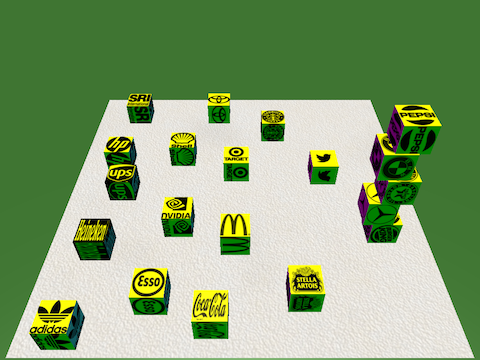}\\
\multicolumn{2}{@{}p{8.1cm}@{}}{\footnotesize place the block that is to the right of the stella block as the highest block on the board. it should be in line with the bottom block .}
\end{tabular}
\caption{Example utterance which requires both understanding that highest is a 3D concept, and inferring that the 2D concept of a line has been rotated to be in the z-dimension.}
\label{tab:3D}
\end{table}

\paragraph{Rotations}
Despite a strong performance by the model on rotations, there are a number of cases that were completely overlooked. Upon inspection, these appear to be predominantly cases where the rotation is not explicitly mentioned, but instead assumed or implied:
\begin{itemize}
\item place toyota on top of sri in the \textbf{same direction} .
\item take toyota and place it on top of sri .
\item ... making part of the inside of the \textbf{curve of the circle} .
\end{itemize}
The first two should be the focus of immediate future work as they only require trusting that a new block should trust the orientation of an existing one below it unless there is a compelling reason (e.g. balance) to rotate it.  The third case, returns to our larger discussion on understanding geometric shapes and is probably out of scope for most approaches.

\section{Conclusions}
This work presents a new model which moves beyond simple spatial offset predictions (+x, +y, +z) to learn functions which can be applied to the scene.  We achieve this without losing interpretability.  In addition, we introduce a new corpus of 10,000 actions and 250,000 tokens which contains a plethora of new concepts (subtle movements, balance, rotation) to advance research in action understanding.

\section*{Acknowledgments}
We thank the anonymous reviewers for their many insightful comments. 
This work was supported in part by the 
NSF grant (IIS-1703166), DARPA CwC program
through ARO (W911NF-15-1-0543), and gifts by
Google and Facebook.

\bibliographystyle{aaai}
\bibliography{references_acro}

\end{document}